\documentclass[letterpaper, 10 pt, conference]{ieeeconf}  % Comment this line out if you need a4paper
\IEEEoverridecommandlockouts                              % This command is only needed if 
\usepackage{graphicx} % for pdf, bitmapped graphics files
\usepackage{amsmath} % assumes amsmath package installed
\usepackage{amssymb}  % assumes amsmath package installed
\graphicspath{{figures/}}

\usepackage{adjustbox}

\usepackage{booktabs} % For prettier tables
\usepackage{siunitx} % scientific notations
\usepackage{hyperref} % for reference urls
\usepackage{multirow}
\usepackage{array} %increases tex's buffer size and enables ``>'' in tablespecs
\usepackage{longtable}
\usepackage{dcolumn} %Aligning numbers by decimal points in table columns
\usepackage{cleveref}

\usepackage{csquotes} % for enquote

\usepackage{optidef}
\usepackage[dvipsnames]{xcolor}
\usepackage{mathtools}
\usepackage{subcaption}
\usepackage[ruled]{algorithm2e}
\SetKwInput{kwInit}{Init}
\SetKwComment{Comment}{$\triangleright$\ }{}
\SetCommentSty{itshape}

\usepackage[]{todonotes}

\usepackage{etoolbox}
\usepackage[noadjust]{cite}

\title{\LARGE \bf
A System for Traded Control Teleoperation of Manipulation Tasks \\
using Intent Prediction from Hand Gestures }

\author{Yoojin Oh$^{1}$\thanks{Corresponding author: \tt{yoojin.oh@ipvs.uni-stuttgart.de}}, Tim Sch\"afer$^{1}$, Benedikt R\"uther$^{1}$, Marc Toussaint$^{2,3}$ and Jim Mainprice$^{1,2}$\\% <-this % stops a space
\authorblockA{$^1$Machine Learning and Robotics Lab, IPVS, University of Stuttgart, Germany}
\authorblockA{$^2$Max Planck Institute for Intelligent Systems ;  MPI-IS ; T{\"u}bingen/Stuttgart, Germany}
\authorblockA{$^3$Technische Universit\"at Berlin ; TUB ; Germany}
}
\vspace{-0.8cm}

\begin{document}
\bstctlcite{IEEEexample:BSTcontrol}

\maketitle
\thispagestyle{empty}
\pagestyle{empty}

%%%%%%%%%%%%%%%%%%%%%%%%%%%%%%%%%%%%%%%%%%%%%%%%%%%%%%%%%%%%%%%%%%%%%%%%%%%%%%%%
\begin{abstract}

This paper presents a teleoperation system that 
includes robot perception and intent prediction from hand gestures.
The perception module identifies the objects present
in the robot workspace and 
the intent prediction module which object the user likely wants to grasp.
This architecture allows the approach to rely on traded control
instead of direct control:
we use hand gestures
to specify the goal objects for a sequential manipulation task,
the robot then autonomously generates
a grasping or a retrieving motion using trajectory optimization.
The perception module relies on the model-based tracker to precisely track
the 6D pose of the objects and makes use of
a state of the art learning-based object detection and segmentation method, 
to initialize the tracker by automatically detecting objects in the scene.
Goal objects are identified from user hand gestures using a trained a multi-layer perceptron classifier.
After presenting all the components of the system and their empirical evaluation,
we present experimental results comparing our pipeline to a direct traded control
approach (i.e., one that does not use prediction)
which shows that using intent prediction allows to 
bring down the overall task execution time.

% Leap Motion \cite{weichert2013analysis}
%This hand gesture controlled user interface,
%tracks accurately hand movements using a depth optical sensor. 
%Traded control refers to the autonomous execution
%of sub-tasks that are specified by the operator. In this paper,
%we present a system that implements such a control strategy by a hand gesture controlled user interface
%controller  as an input device.
%While this type of interface is quite intuitive, it

\end{abstract}

%%%%%%%%%%%%%%%%%%%%%%%%%%%%%%%%%%%%%%%%%%%%%%%%%%%%%%%%%%%%%%%%%%%%%%%%%%%%%%%%
\section{Introduction}

Intelligent robots can substitute or assist humans to accomplish
complicated and laborious tasks. 
They are becoming present in our lives from production lines to hospitals
and our homes. However, many applications remain challenging for
robots to function in full autonomy. Teleoperation is an intermediate solution for controlling robots in scenarios where the task objectives have to be
decided in real-time, such as disaster relief \cite{phillips2016autonomy},
autonomous driving \cite{johns2016exploring}, or assistive devices~\cite{muelling2017autonomy,goil2013using}.

\textit{Shared control} has been investigated to
effectively blend user and autonomous control during teleoperation.
The linear blending paradigm introduced by Dragan et. al~\cite{dragan2013policy}
is still widely applied 
in many shared control frameworks~\cite{goil2013using,anderson2014experimental,gao2014contextual}. 
In the approach, the amount of arbitration is dependent on the confidence of user prediction.
However, the user loses control authority when the robot predicts the user's intent with high confidence.

Some works allocate maximum control authority to the user
by providing minimal assistance only when it is necessary.
Broad et. al.~\cite{broad2018operation} introduced
minimum intervention shared control that computes whether the control signal 
leads to an unsafe state and replaces the user control if so.
Our recent work~\cite{oh2020natural}
formulates shared control as an optimization problem,
which can conveniently balance control authority and optimality
when a complete robot policy is available.
%The shared action is chosen to maximize
%the user's internal action-value function while constraining the
%shared control policy to not diverge away from the autonomous robot policy.

While these works are relevant to the teleoperation of simple manipulation
task where direct control is not optimal,
they are generally limited to controlling the end-effector of the robot
by blending between direct and autonomous control.
They rely on a semantic mapping of the workspace
but they do not let the autonomy take complete advantage of these models,
so as to maximize control authority.
Additionally when using interfaces such as hand gestures controllers,
direct teleoperation is often nearly impossible as the mismatch between
the kinematics of the robot and hand gestures is too large to produce fluid movements.

\begin{figure}[t]
\centering
\includegraphics[clip,trim={3.5cm 1.2cm 1cm 1.3cm},width=0.50\textwidth]{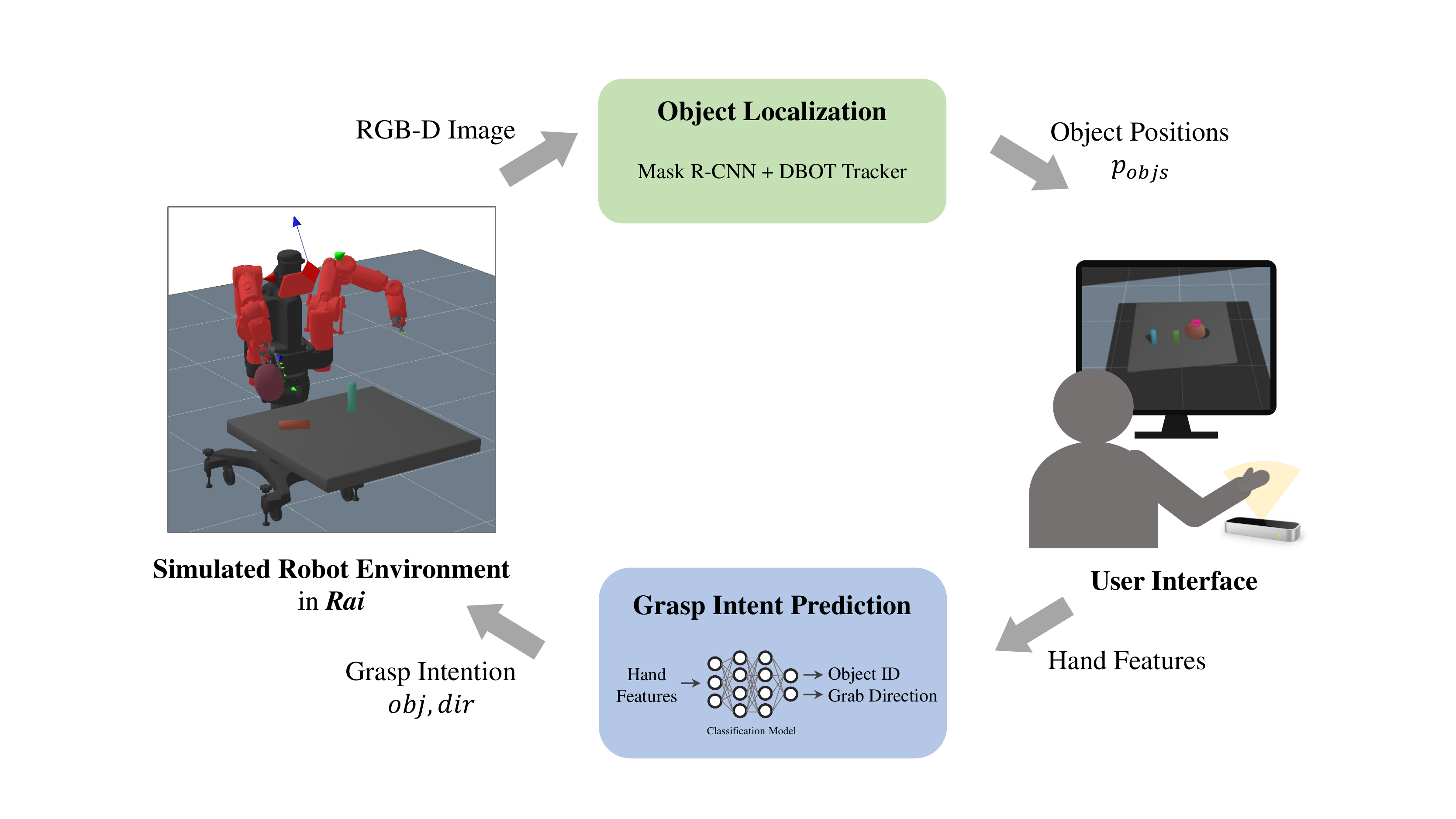}
\caption{Overview of the teleoperation system}
\label{fig:pred_res}
\vspace{-.3cm}
\end{figure}

Hence in this systems paper, we demonstrate
a complete \textit{traded control} teleoperation implementation,
where the user specifies the task objectives and executes the motion autonomously.
Contrarily to the aforementioned approaches, it does not blend between direct and autonomous control.
Our system makes use of available models,
in terms of object poses and shapes to plan robot motion trajectories.

We present and evaluate all components needed for such a system,
which can be decomposed into three parts: 
1) a perception pipeline
capable of identifying and tracking objects,
2) an intent estimation system that can identify which objects
to grab and how, 3) a motion planning system that can produce accurate
manipulation motion in accordance to the human intent.
\footnote{Video available at \url{https://sites.google.com/view/ohyn-teleoperation-pipeline/home}}

%% add advantages
%Our approach allows the user to maintain more control authority over the system compared to directly blending the policies. In addition, the method can provide different levels of assistance depending on the cost function that the robot policy optimizes. That is, the framework can be flexibly applied independent of the presence of a goal. 

After presenting the individual components,
we assess the accuracy of the 
the different modules on dedicated tasks.
We evaluate the object localization and tracking 
module on several objects in simulation, test the accuracy of our grasp intent prediction module using a dataset of trajectories.
Finally, we present results using our grasp intent inference module,
where various users are simulated using
degraded user trajectories collected using the hand gesture controller.

We summarize our main contributions as the following:
\begin{itemize}
	\item A teleoperation system capable of traded control using hand gestures
		\item Simulated user experiment assessing the capacity of
	our grasp intent prediction module to perform teleoperation of pick and place motions
	\item A solution for automatic initialization of an existing object tracking module using Mask R-CNN \cite{he2017mask}

\end{itemize}

This paper is structured as follows: we present related work in
Section~\ref{sec:rel_work}.

Section~\ref{sec:user_interface} presents our user interface.
Section~\ref{sec:tracking} presents our object tracking pipeline, including
the combination of Mask R-CNN and a model-based object tracker.
Section~\ref{sec:experiments} presents the assessment of the modules in our pipeline.
Conclusions are drawn in Section~\ref{sec:conclusions}.

%This has been reported to generate mixed preferences from users
%where some users prefer to keep control authority despite longer
%completion times~\cite{kim2011autonomy, Javdani:2018bt}.
%Additionally, when assistance is against the user's intentions
%this approach can aggravate the user's workload~\cite{dragan2013policy};
%the user ``fights" against the assistance rather than gain help from it.

%%%%%%%%%%%%%%%%%%%%%%%%%%%%%%%%%%%%%%%%%%%%%%%%%%%%%%%%%%%%%%%%%%%%%%%%%%%%%%%%
\section{Background and Related Work}
\label{sec:rel_work}

%Prior work in the area of disaster recovery robotics \cite{Yanco:04} has revealed the need for better group organization, perceptual and assistive interfaces, models of the state of the robot and its surroundings, and information as to what has been observed. 
%In this work, we present a solution for a teleoperation system using traded control that combines previous works on object localization and tracking.

\subsection{Traded Control in Teleoperation}

Traded control is a discrete switching mechanism between high-level robot autonomy and low-level control depending on predefined circumstances. It is also referred to as \textit{control switching}, as the system allocates all-or-none assistance rather than a blended spectrum between user and robot controls. 
%It can be momentary: at each state the robot evaluates whether to take over depending on the confidence of the user's intentions~\cite{dragan2013policy}~(aggressive mode); or to prevent the system from entering an unsafe state~\cite{broad2018operation}. 
The operator initiates a sub-task or behavior for the robot
and the robot performs the sub-task autonomously while the operator monitors the robot~\cite{kofman2005teleoperation, phillips2016autonomy}. \cite{bohren2017preliminary} showed that intent-based traded control can improve teleoperation performance and alleviate difficulties in high-latency teleoperation scenarios.

\subsection{Hand Gesture Recognition for Robot Control}

The Leap Motion controller (Ultra Leap, https://www.ultraleap.com/) is a consumer-grade, marker-less motion capture sensor that tracks hand gestures and finger movements up to 200 Hz. 
\cite{weichert2013analysis} showed that its accuracy is below 2.5mm, however, the controller shows inconsistent performance due to its limited sensory range ~\cite{guna2014analysis}. Nevertheless, its simplicity and its capability to track the hand in 6-Dof are the reasons for its application.

Prior works used deep learning to improve the accuracy of the gesture recognition, such as SVMs and random forests~\cite{marin2016hand}, or neural networks using radial basis functions (RBF)~\cite{zeng2018hand}. Similar to~\cite{qi2021multi}, we propose to train a gesture classifier (i.e., which object
is intended) for hand motion recognition rather than mapping hand
features directly to robot configurations.
Achieving higher accuracy is easier on classification than regression
(i.e., predicting accurate positions) tasks,
which is one of the justifications for our traded control approach.

%\subsection{Service-Oriented Architectures}
%
%The framework we present has been developed within a Service-Oriented Architecture (SOA) which consists of distinct software modules that communicate with each other \cite{Jackson:07,HAMMER_2011,Quigley_2009}. We chose ROS \cite{Quigley_2009} for its ease-of-use and visualization tool (RViz).
%SOAs have become a popular choice for robotics since they allow the software to be highly modular and adaptive \cite{Pordel_2013}.
%There are many options for SOAs currently available today \cite{Jackson:07,HAMMER_2011,Quigley_2009}. We have chosen ROS to implement this framework due to its extensive proven ability to control high-DoF robots such as the PR2. ROS has also been applied to more anthropomorphic humanoid robots such as Nao \cite{Almetwally_2013}.
%Robonaut 2, %\cite{Diftler_2011}, 
%and TU/e TUlip, % \cite{Hobbelen_2008}.
%Additionally, ROS was chosen for its built-in visualization tool (RViz) which allows for fast operator interface development. For further surveys on robotic frameworks, both free and commercially available, see \cite{Harris_2011, Craighead_2007}.

\subsection{Depth Based Object Tracking (DBOT)}

%6D Pose estimation and object tracking is a intensely investigated 
%field of robot manipulation \cite{collet2011moped}.
%Especially, the advances in high performance computing and the use of convolutional neural networks lead to significantly better detection/prediction performance \cite{sermanet2014overfeat, xiang2018posecnn, schwarz2015rgbd}.
%
%Traditional approaches use template-based methods, which obtain the pose by rendering the objects from different fixed camera poses and try to find the best matching template all across the image \cite{su2015render, rad2018bb8}. These methods with templates are simple and able to detect texture-less objects but often fail with bad lighting conditions or occlusions between the objects.
%By taking the depth information into account, the performance in bad lighting or partial occlusion increases \cite{schwarz2015rgbd}.
We utilize the implementation of depth-based object tracking methods described in~\cite{wuthrich2013probabilistic} (``particle tracker'') and~\cite{issac2016depth} (``Gaussian tracker'') to acquire the 6D pose of objects during teleoperation. Compared to recent learning-based methods such as PoseCNN \cite{xiang2017posecnn} and DenseFusion \cite{wang2019densefusion}, the methods take a model-based approach.

The particle tracker in DBOT tracks objects by computing a posterior distribution over the object using a dynamic Bayesian network for inference~\cite{wuthrich2013probabilistic}, while the Gaussian tracker improves the performance of a Gaussian filter using a robustification method as well as reducing the filter's computational complexity~\cite{issac2016depth}.
This approach has the advantage of being robust
without requiring any extra tuning or pre-training.

%%%%%%%%%%%%%%%%%%%%%%%%%%%%%%%%%%%%%%%%%%%%%%%%%%%%%%%%%%%%%%%%%%%%%%%%%%%%%%%%

\section{Teleoperation using Traded Control}
\label{sec:user_interface}
\subsection{Hand Gesture Based Robot Control}
The user provides grasping intentions and commands by performing reach-and-grasp motions with the right hand as if the user naturally reaches and grasps an object while looking at the environment from the robot's perspective.

\begin{figure}[h]
	\hspace{-.3cm}
     \centering
     \begin{subfigure}{0.33\linewidth}
         \centering
         \includegraphics[clip,trim={2cm 0cm 4cm 0cm},width=\textwidth]{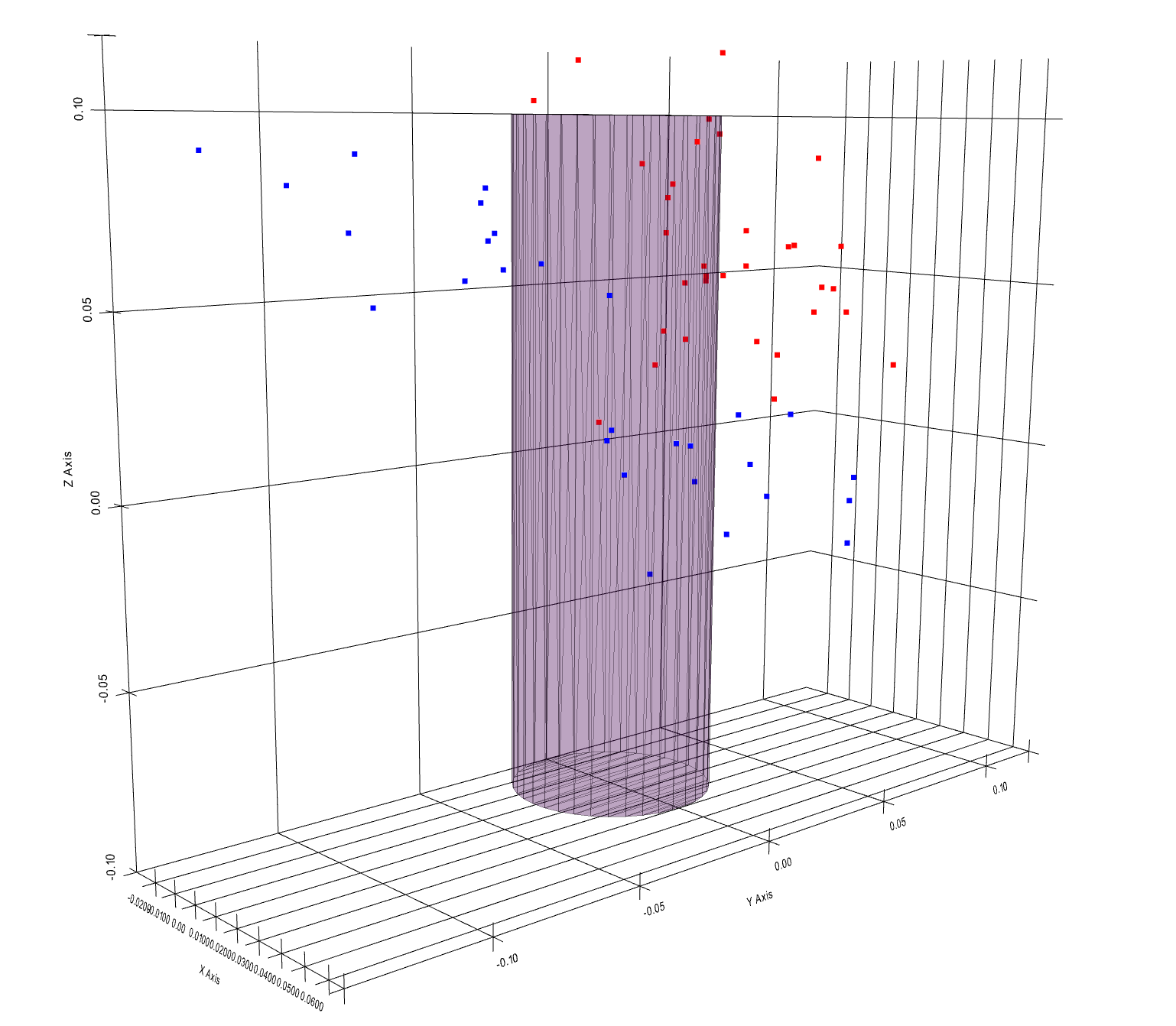}
                                            \caption{Grab positions}
                                          \label{fig:grab_positions}
     \end{subfigure}     
     \hfill
     \begin{subfigure}{0.315\linewidth}
         \centering
         \includegraphics[clip,trim={10cm 8cm 10cm 7cm},width=\textwidth]{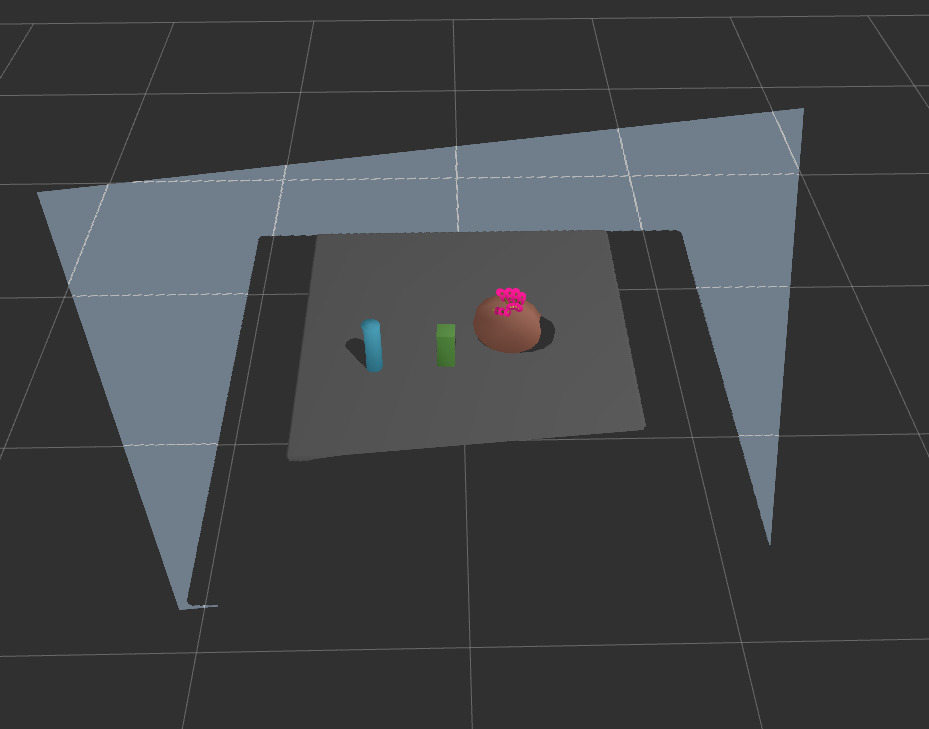}
                                            \caption{Disk top grab}
                                          \label{fig:top_grab}
     \end{subfigure}
     \hfill
 	 \begin{subfigure}{0.33\linewidth}
         \centering
         \includegraphics[clip,trim={10cm 8cm 10cm 7cm},width=\textwidth]{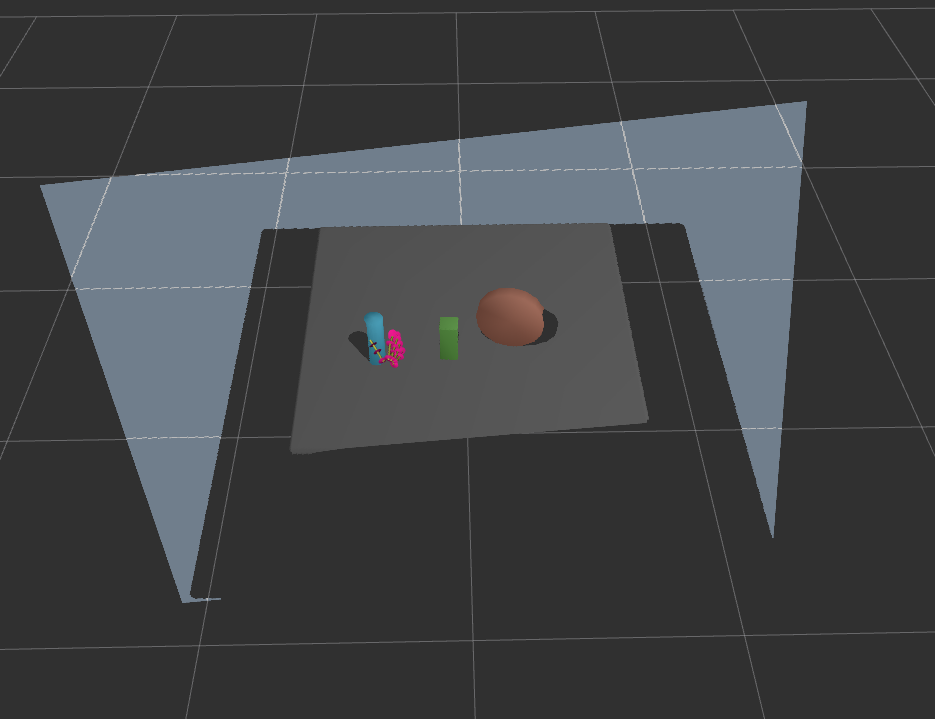}
                                            \caption{Cylinder right grab}
                                          \label{fig:right_grab}
     \end{subfigure}
     \caption{(a) Grab positions from users tracked with the Leap Motion controller, top grab (red)/right grab (blue). (b), (c) User interface for reach and grab motion in a setting with three objects.}
     \label{tab:grab_settings}
     \vspace{-.5cm}
\end{figure}

The hand motion is captured using a Leap Motion controller. 
Features are captured and published via ROS topics at a frame rate of 180Hz.
Since the user is actually reaching towards an invisible object, the grabbing positions vary significantly as shown in~\ref{fig:grab_positions}. We resolve this issue with a traded control paradigm and learning a classifier to distinguish how the user is intending to grab the object.
%such as the palm position, hand direction vector, the normal vector to the palm, and roll/pitch/yaw rotation of the hand 

\subsection{Traded Control}

To alleviate the inconsistent hand tracking performance, 
%while taking advantage of an unconstrained hand control, 
we adopt a traded control method rather than a continuous shared control paradigm. This also relieves the problems that arise from the physical difference between the human arm and the robot arm.

Once objects are identified as described in Section~\ref{sec:tracking}, we predict the user's intent of the target object and in which direction the user is intending to grasp. As soon as the intention is identified, the robot controller executes the object reach and grab motion. The user still maintains control authority by having the ability to decide in which order to grab the set of objects, that is, we rely on the human user for high-level decision making and the robot takes care of the low-level control and motion planning. 
%Knowing in advance which object and how the user wants grasp the user wants the robot to grasp is advantageous in planning the robot grasp motion. 

\subsection{Grasp Intention Prediction}

We train a multi-layer perceptron using supervised learning to classify the goal object and the grasp direction. We assume a fixed set of objects ($m$=3) along with their positions and two possible grab directions (top/right, $n$=2), as shown in Figures~\ref{fig:top_grab} and \ref{fig:right_grab}.

The input includes eight features: distances from the hand objects, x-component of the hand position, x-component of the hand direction, x,y-components of the palm normal vector, and y-rotation of the hand are selected through experience. The model consists of three dense layers of 64 hidden units that are connected to two separate layers of two units and outputs the class labels. 

%%%%%%%%%%%%%%%%%%%%%%%%%%%%%%%%%%%%%%%%%%%%%%%%%%%%%%%%%%%%%%%%%%%%%%%%%%%%%%%%

\begin{figure}[t]
  \centering
  \includegraphics[clip,trim={0.04cm 0cm 0.04cm 0cm},width=0.34\textwidth]{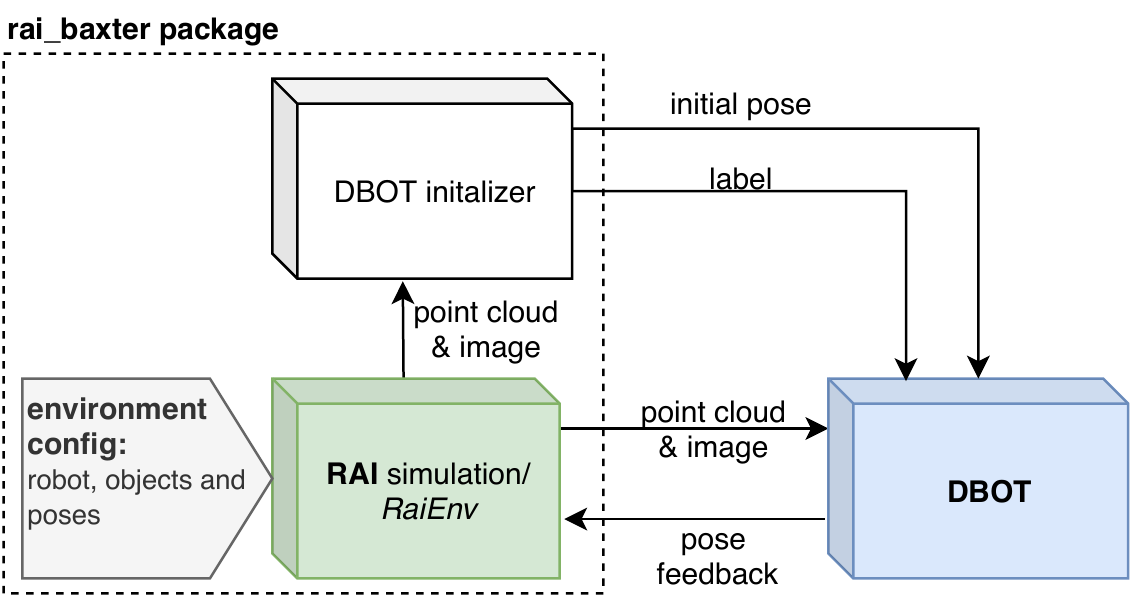}
  \includegraphics[clip,trim={0cm 0cm 0cm 2cm},width=0.13\textwidth]{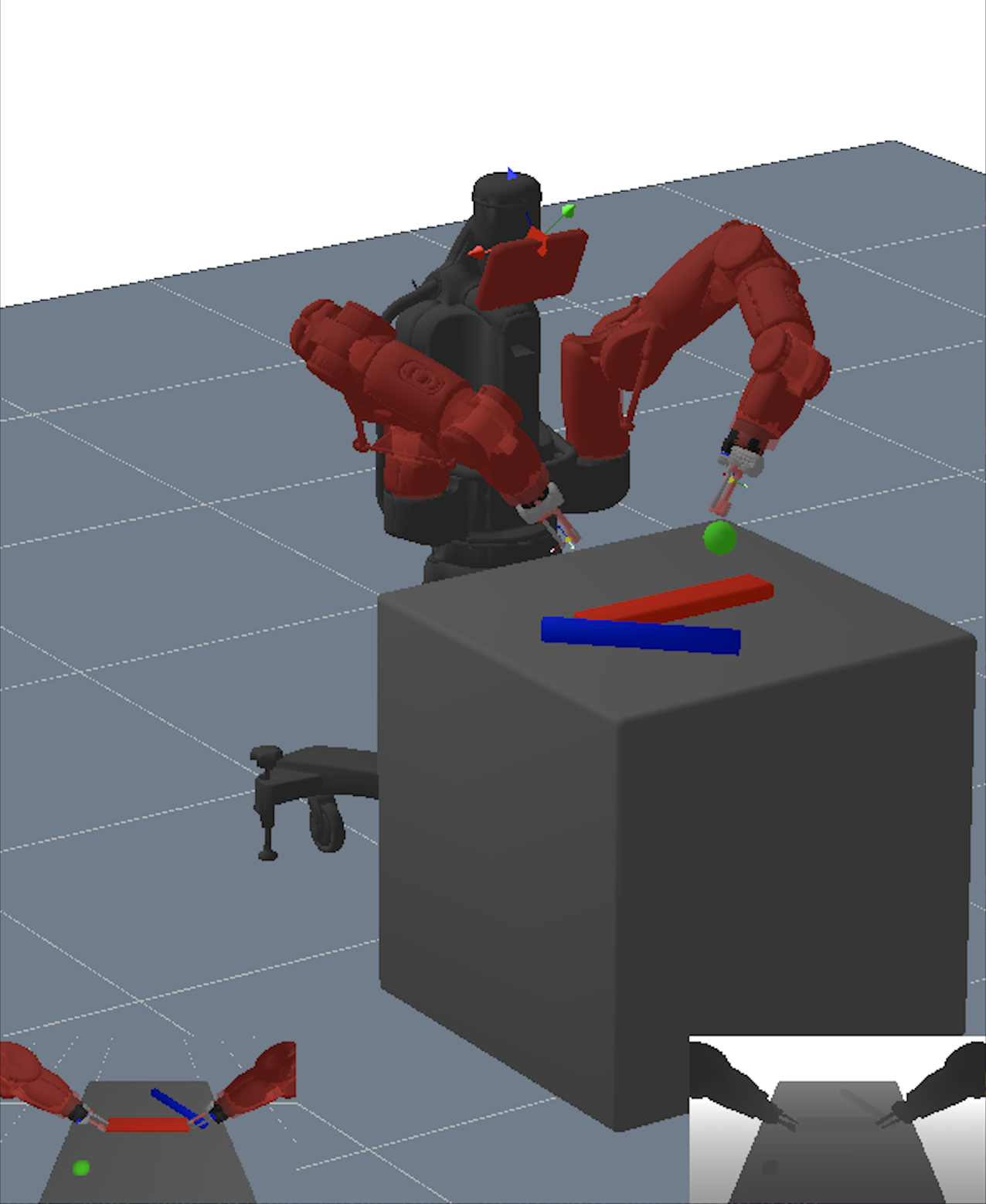}
  \caption{Overview of the object tracking pipeline and an image of the simulated robot environment.}
  \label{fig:rai_interface}
 \vspace{-.5cm}
\end{figure}

\section{Object Tracking Pipeline}
\label{sec:tracking}
\begin{figure}[t]
\centering
\includegraphics[width=.65\columnwidth]{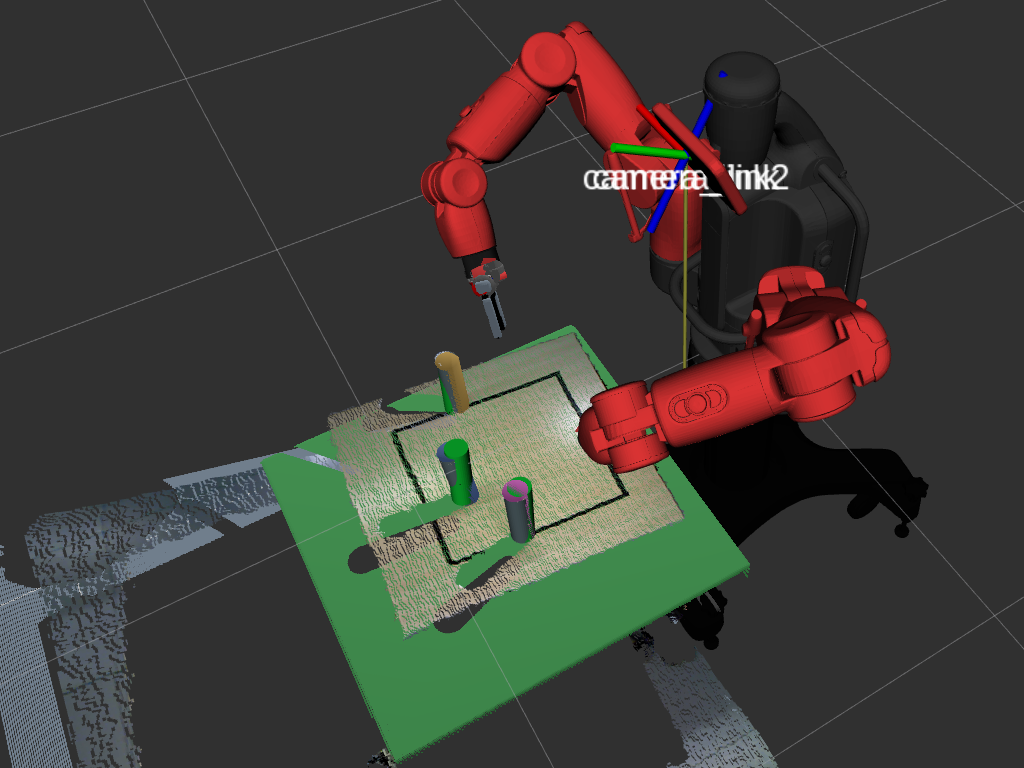}
\caption{Snapshot of our object tracking with real images}
\label{fig:baxter}
\vspace{-.5cm}
\end{figure}

The object tracking pipeline is automatically initialized and tracks the 6D pose of rigid objects. We make use of the existing object tracking library \cite{wuthrich2013probabilistic,issac2016depth} and provide a solution to alleviate the burden of manual initialization.
The pipeline consists of two modules:
%the simulated environment (RaiEnv),
object tracker initializer (DBOT initializer) and the object tracker (DBOT) as shown in Figure~\ref{fig:rai_interface}.
The DBOT initializer receives camera images of the environment (see Figure \ref{fig:baxter}) and predicts the initial pose as well as the semantic label of observed objects. 
%The DBOT tracker continuously tracks the objects given the initial pose and label of the object. The following sections describe each component in detail.

%First starting with the simulation (Section 3.2), where the robot and the surrounding environment is
%simulated with a physics engine. Sensor and camera output of the robot are periodically published
%as ROS topics, which can be subscribed by RViz or DBOT. Secondly, instance segmentation
%(Section 3.3) and pose estimation (Section 3.4) are providing the DBOT initialization part of
%the pipeline, which is a major part in the thesis. Mask R-CNN [Abd17] is used for the instance
%segmentation. The masks are the input for the pose estimation. This pose estimation is a set of
%methods and algorithms to estimate the pose of the objects, from the mask and the depth information.
%Finally DBOT, which is described in Section 2.4.2, receives these initial poses. After receiving the pose the tracker starts working by collecting the point cloud updates from the simulation via ROS
%topic subscription.

\subsection{Automated Object Tracker Initialization}
\label{ssec:pose_init}

DBOT assumes that the initial pose of the object is given.
In practice, this initialization is done by the user by manually positioning a marker over the object's depth image in a
3D visualization of the depth image.
We use Mask R-CNN \cite{he2017mask}, a state of the art instance segmentation method, to automate the initialization process by identifying the masks and the labels using transfer learning.

%The mask is used to segment the depth image and obtain depth pixels that correspond to the object. The depth pixels are converted into a 3D point cloud by applying pixel-to-point transformation using the pinhole camera model. 
%Outliers in the boundary of the point cloud are removed.

Once the depth pixels are segmented from the depth image using the object mask, we use point cloud registration to compute the object's 6D pose.
Utilizing the labels from the Mask R-CNN, the corresponding mesh model of the object is loaded and a set of points are sampled from the mesh as a reference point cloud for registration.
Rigid registration is performed using the coherent-point-drift (CPD) algorithm~\cite{probreg}, and we use the average position of the masked point cloud as a rough initialization during the registration. The output of the CPD is a 4$\times$4 homogeneous transformation matrix. We refer to this estimated pose as \textit{mesh\_pose}. 

The process takes approximately 1-16 seconds, depending on the number of tracked objects and the number of iterations during point cloud registration. The step is performed once to initialize the DBOT tracker.
% and we rely on DBOT for pose estimation when the tracker is running. 

\subsection{DBOT Tracker}
\label{ssec:DBOT}

%DBOT\cite{dbot_code} requires manual initialization in RViz, where an interactive maker has to be manually aligned to the point cloud. To initialize DBOT  %according to the simulated environment 
%without RViz, DBOT provides a ROS service, which receives the initial pose for the particle tracker.
%Additionally, a new ROS service for the Gaussian tracker is implemented to initialize this tracker as well. 
%
%Both services are extended to receive initialization for multiple objects and to track them in parallel. A call to initialize the services contains a list of individual names, mesh-filenames (3D models) and poses. DBOT requires the object poses relative to the camera coordinates. 

%After the initialization, the DBOT trackers compute a new position for every object on each update of the point cloud, via the ROS publisher of the simulation.  % (see \cref{fig:rai_baxter}).
%In RViz, the tracking can be visualized by adding markers as shown in \cref{fig:rviz_mult_mark}.
%The DBOT tracker is initialized automatically using this pipeline.
%The tracker is initiated using ROS services. A call to initialize the services contains a list of individual names, mesh-filenames (3D models) and poses. 
%DBOT requires the object poses relative to the camera coordinates. 
%The services are extended to receive initialization for multiple objects and to track them in parallel.

Once the DBOT trackers are successfully initialized, the simulation or a real robot system can subscribe to the refined poses from DBOT and use the information to perform precise object manipulation.
The DBOT tracker runs with 10Hz for up to 7 objects on CPU. The performance can be further improved by utilizing GPU.

\begin{figure*}[]
\centering
\begin{minipage}[b]{0.40\linewidth}
     \begin{subfigure}{0.20\textwidth}
         \centering
         \includegraphics[width=\textwidth]{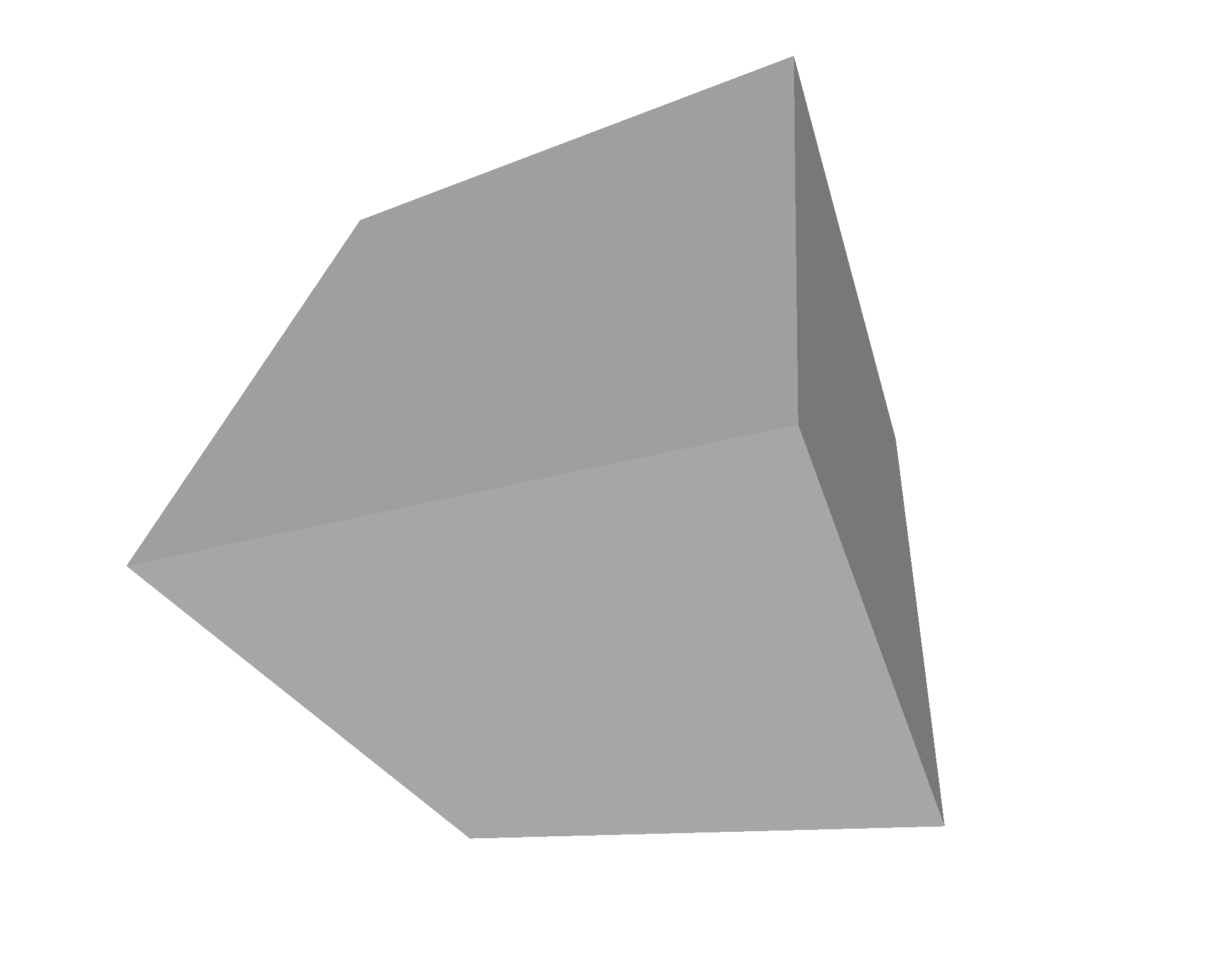}
                                            \caption{cube}
                                          \label{fig:cube}
     \end{subfigure}
     \hfill
     \begin{subfigure}{0.20\textwidth}
         \centering
         \includegraphics[width=\textwidth]{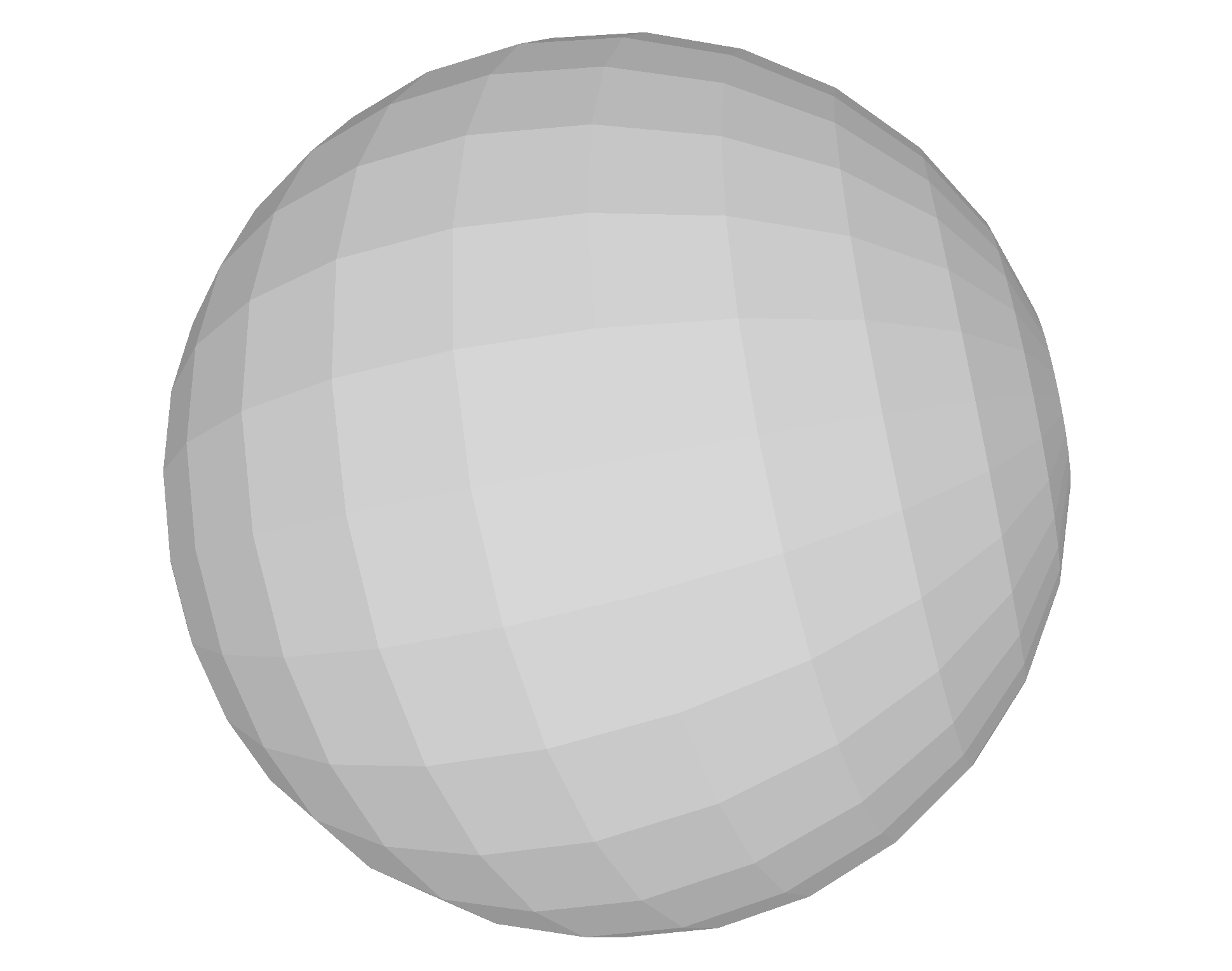}
                                            \caption{sphere}
                                          \label{fig:sphere}
     \end{subfigure}
     \hfill
     \begin{subfigure}{0.21\textwidth}
         \centering
         \includegraphics[width=\textwidth]{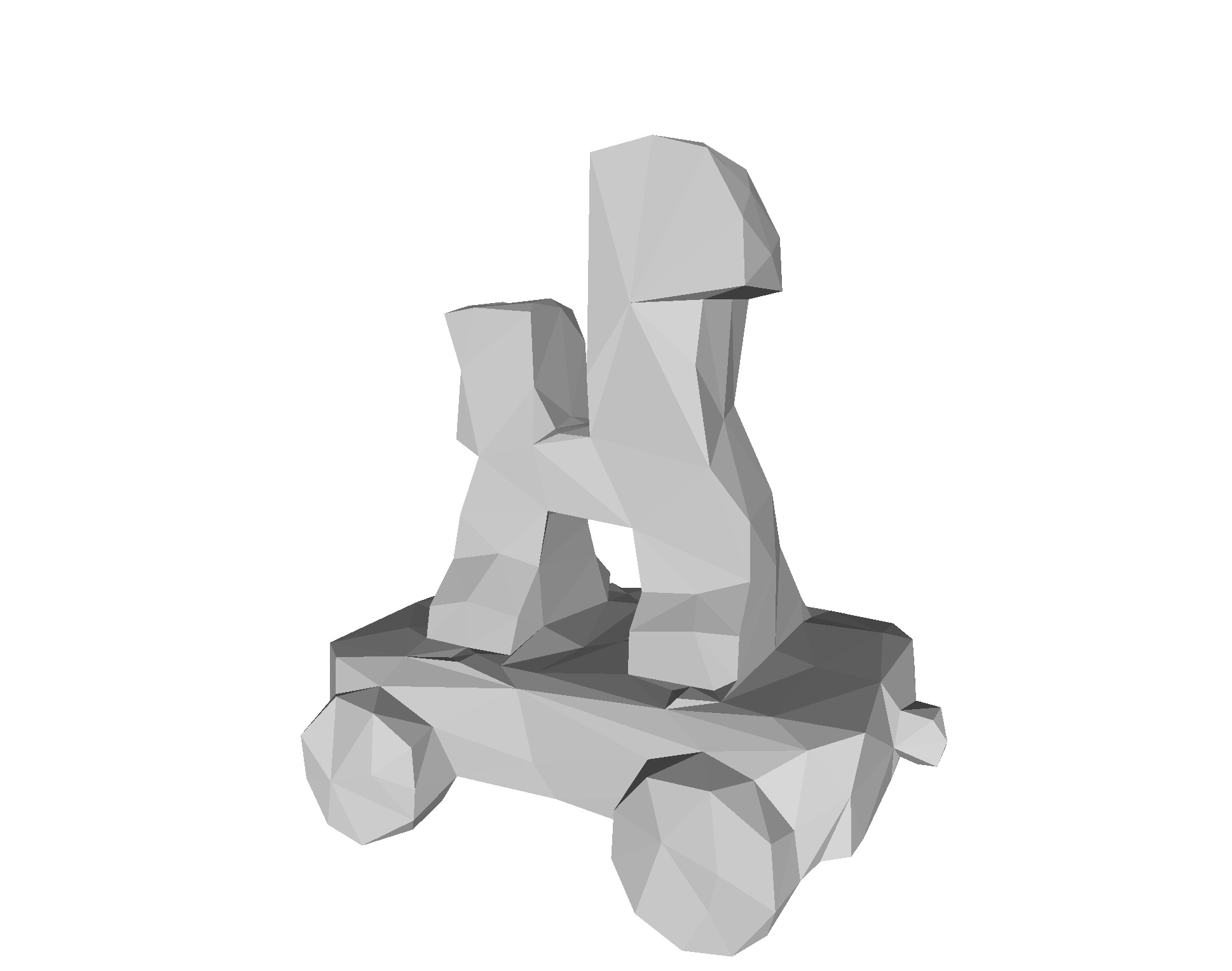}
                                            \caption{lego toy}
                                          \label{fig:lego_toy}
     \end{subfigure}
     \hfill
     \newline
     \begin{subfigure}{0.20\textwidth}
         \centering
         \includegraphics[width=\textwidth]{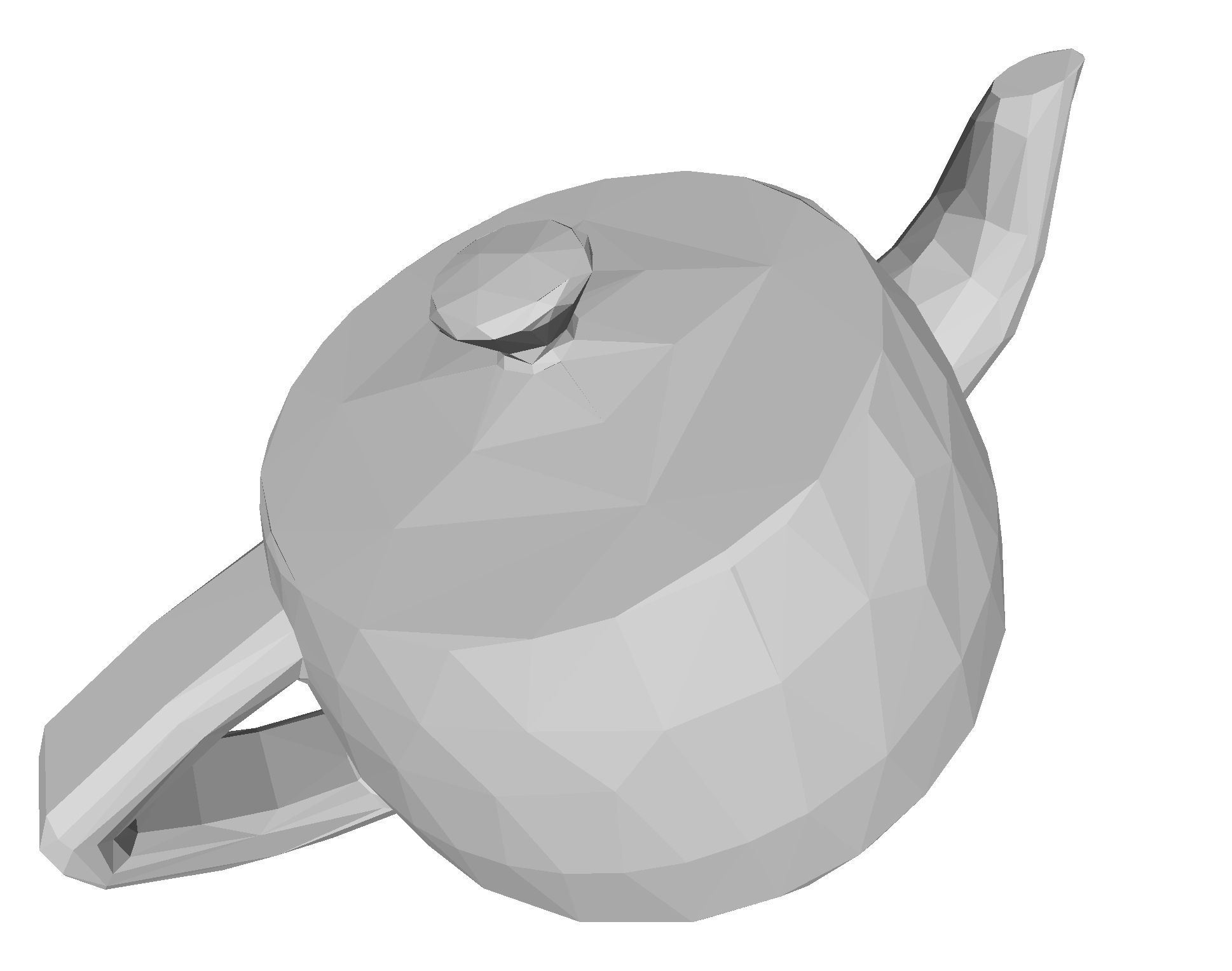}
                                            \caption{teapot}
                                          \label{fig:teapot}
     \end{subfigure}
     \hfill
     \begin{subfigure}{0.20\textwidth}
         \centering
         \includegraphics[width=\textwidth]{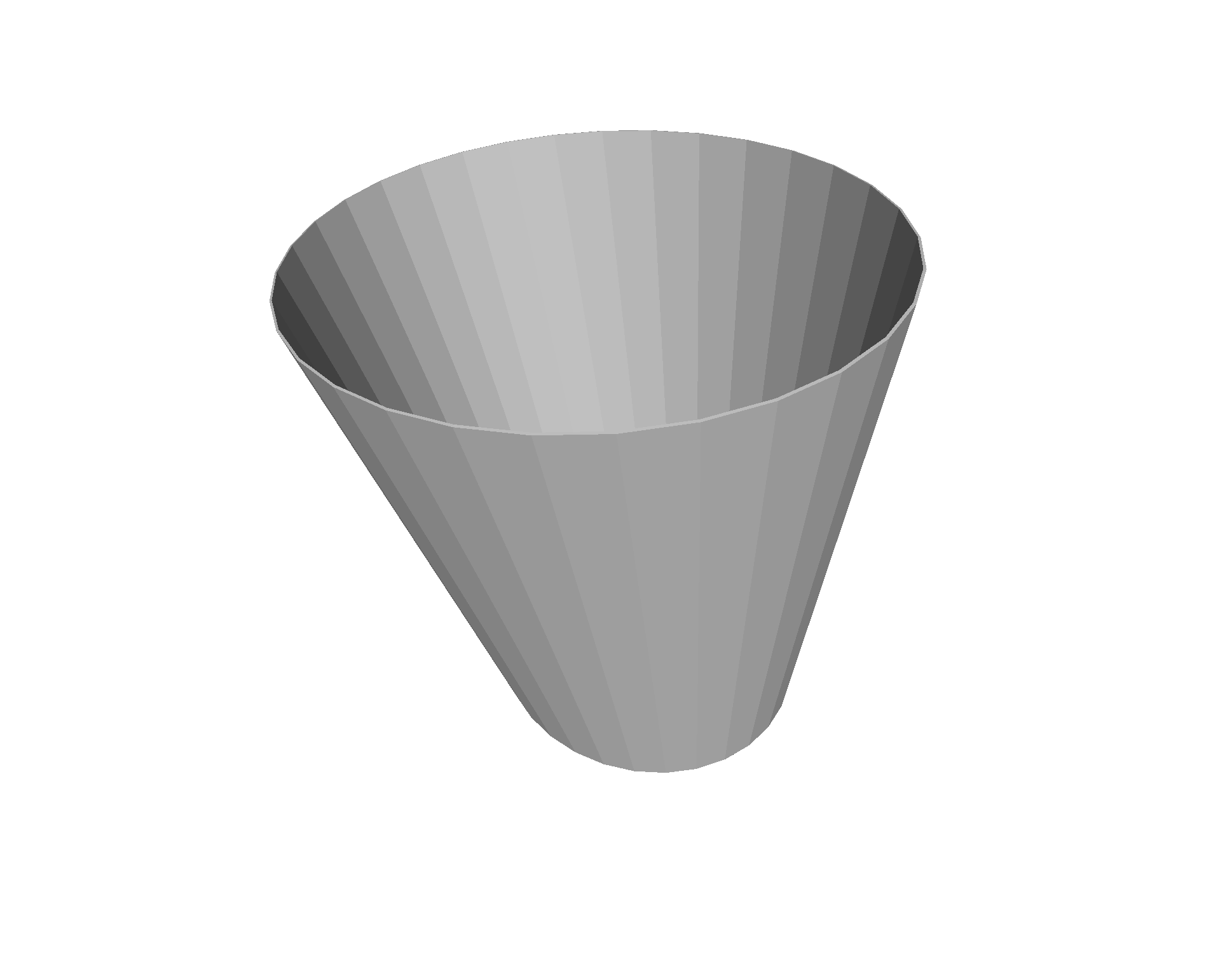}
                                            \caption{cup}
                                          \label{fig:cup}
     \end{subfigure}
     \hfill
     \begin{subfigure}{0.20\textwidth}
         \centering
         \includegraphics[width=\textwidth]{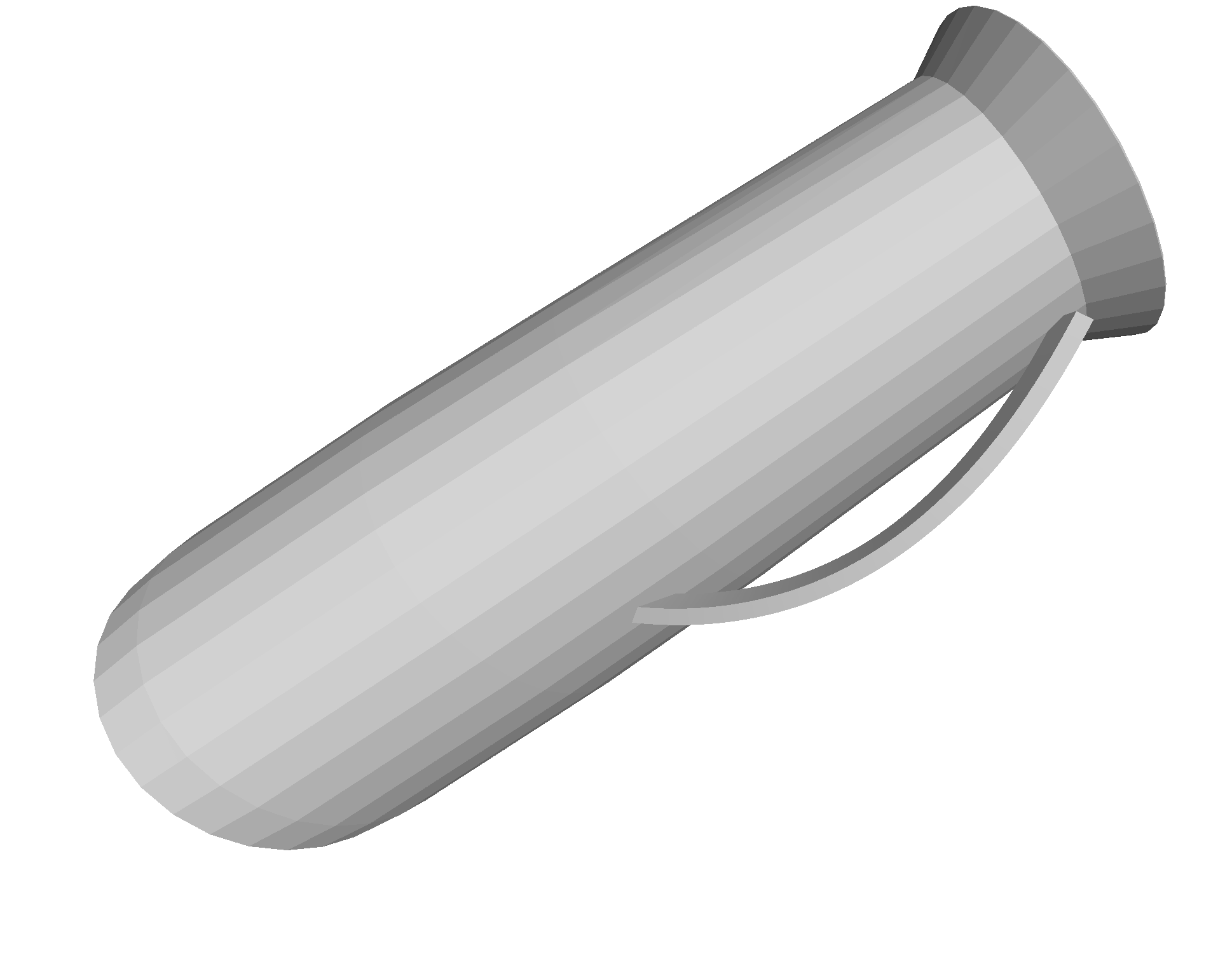}
                                            \caption{jug}
                                          \label{fig:jug}
     \end{subfigure}
     \hfill
     \begin{subfigure}{0.20\textwidth}
         \centering
         \includegraphics[width=\textwidth]{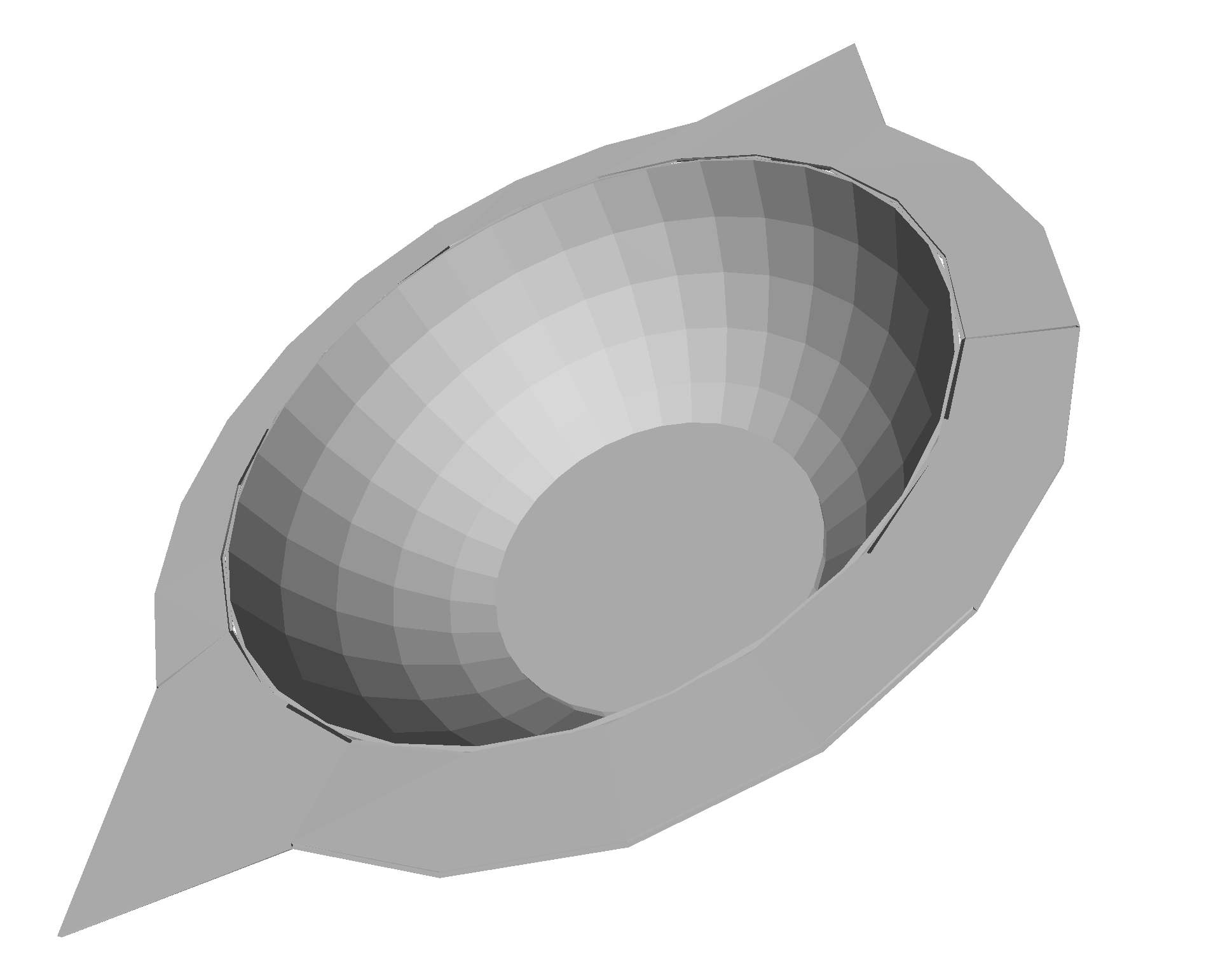}
                                            \caption{bowl}
                                          \label{fig:bowl}
     \end{subfigure}         
\captionof{figure}[objects]{Objects used to train Mask R-CNN}
\label{fig:dataset_objects}     
     \hfill
\includegraphics[clip,trim={0.5cm 0.3cm 0cm 0.5cm},width=1.0\textwidth]{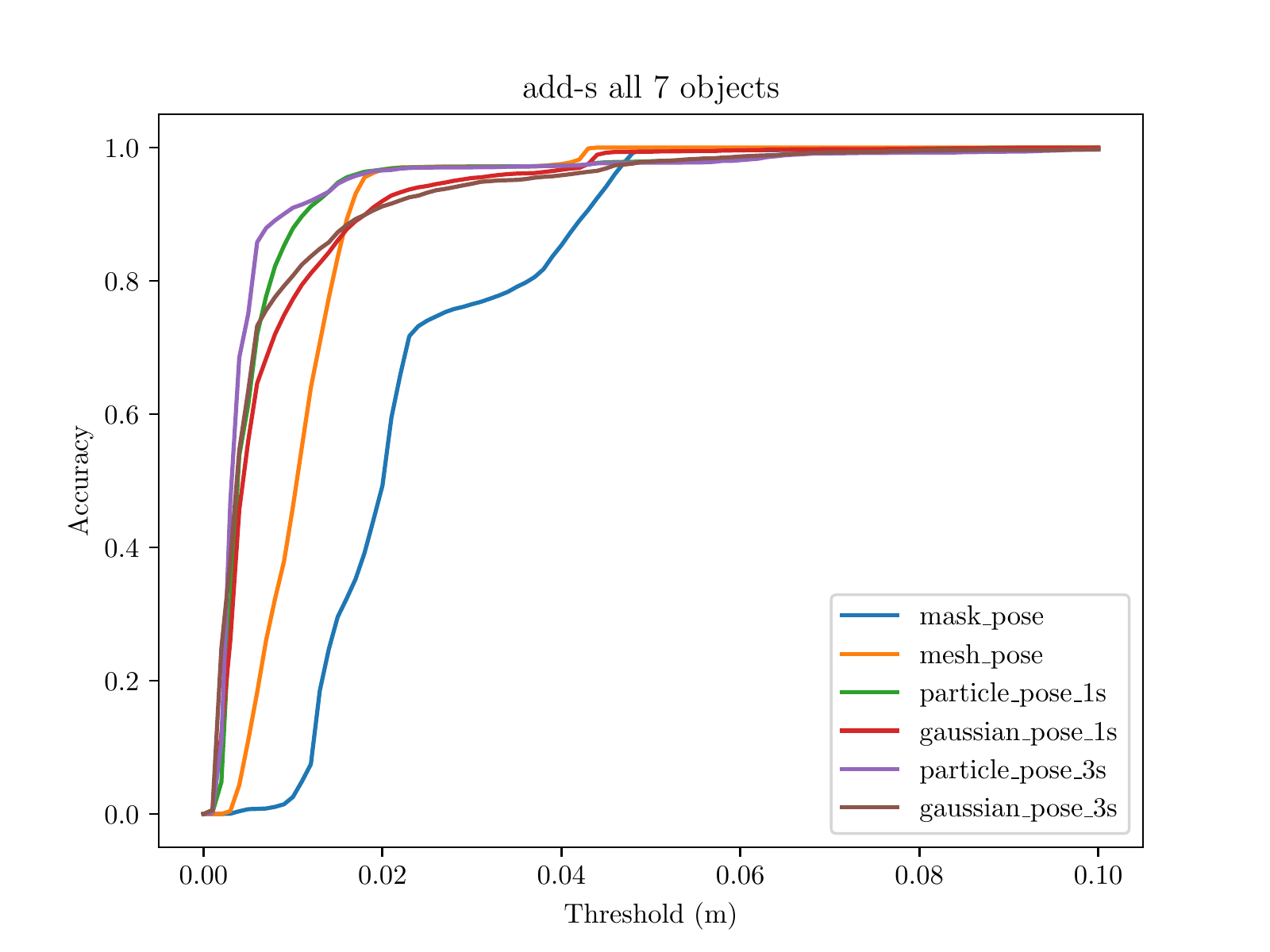}
\vspace{-0.5cm}
\captionof{figure}[Accuracy-Threshold Curve]{Accuracy-threshold curve of the average ADD-S for seven objects with maximum threshold of 10cm.}

\label{fig:acc-thres_curve}
\end{minipage}\hfill
\begin{minipage}[b]{0.58\linewidth}
\scriptsize
\setlength\tabcolsep{5pt} % default value: 6pt

\begin{tabular}{lrrrrrrrrrrrr}
\toprule
{} & \multicolumn{1}{l}{mask} & \multicolumn{2}{l}{mesh} & \multicolumn{2}{l}{particle 1s} & \multicolumn{2}{l}{gaussian 1s}     & \multicolumn{2}{l}{particle 3s} & \multicolumn{2}{l}{gaussian 3s} \\
{}          &    t &    t &    R &           t &    R &           t &    R &           t &    R &           t &    R \\
\midrule
cube        & 89.1 & 90.3 & 96.0 &        97.4 & 97.0 &        93.9 & 97.0 &        \textbf{97.7} & \textbf{97.2} &        90.7 & 96.9 \\
sphere      & 76.1 & 79.2 & \textbf{97.0} &        97.9 & \textbf{97.0} &        97.5 & \textbf{97.0} &        97.9 & \textbf{97.0} &        \textbf{98.0} & \textbf{97.0} \\
lego\_toy   & 66.7 & 80.2 & 90.2 &        \textbf{85.0} & 92.2 &        84.8 & 92.3 &        84.9 & \textbf{92.4} &        83.0 & 92.3 \\
teapot      & 77.8 & 80.0 & 90.7 &        91.7 & 93.3 &        86.5 & 92.8 &        \textbf{94.0 }& \textbf{95.2 }&        91.2 & 94.8 \\
cup         & 90.9 & 90.2 & 95.0 &        94.0 & 95.9 &        85.1 & 95.6 &        \textbf{94.3} & \textbf{96.2} &        77.1 & 95.7 \\
jug         & 69.8 & 77.3 & 90.2 &        93.6 & 92.4 &        74.8 & 92.8 &        \textbf{95.1} & 95.3 &        89.9 &\textbf{ 96.2} \\
bowl        & 69.3 & 70.4 & 84.4 &        81.0 & 87.7 &        \textbf{86.6} & 84.9 &        81.1 & \textbf{88.7} &        80.8 & 85.6 \\
\midrule
MEAN        & 77.1 & 81.1 & 91.9 &        91.5 & 93.7 &        87.0 & 93.2 &        \textbf{92.1} & \textbf{94.6} &        87.2 & 94.1 \\
\bottomrule
\end{tabular}
\vspace{1cm}
\centering
\setlength\tabcolsep{5pt} % default value: 6pt
\begin{tabular}{lrrrrrrrrrrrr}
\toprule
{} & \multicolumn{2}{l}{mesh} & \multicolumn{2}{l}{particle 1s} & \multicolumn{2}{l}{gaussian 1s}     & \multicolumn{2}{l}{particle 3s} & \multicolumn{2}{l}{gaussian 3s} \\
{}          &  AUC &  $<$2cm &         AUC &  $<$2cm &         AUC &  $<$2cm &         AUC &  $<$2cm &         AUC &  $<$2cm \\
\midrule
cube        & 93.8 & 100.0 &        96.8 & 100.0 &        95.6 &  99.7 &        \textbf{97.0} & 100.0 &        94.3 &      98.8 \\
sphere      & 88.6 & 100.0 &        \textbf{96.1} & 100.0 &        96.0 & 100.0 &        \textbf{96.1} & 100.0 &        \textbf{96.1} &     100.0 \\
lego\_toy   & 90.8 &  99.3 &        93.3 &  98.1 &        \textbf{93.4} &  96.7 &        \textbf{93.4} &  97.3 &        92.3 &      88.4 \\
teapot      & 88.2 & 100.0 &        93.7 & 100.0 &        91.3 &  94.0 &        \textbf{95.2} & 100.0 &        93.8 &      95.2 \\
cup         & 92.9 & 100.0 &        95.1 &  99.0 &        92.0 &  92.7 &        \textbf{95.3} &  98.9 &        87.8 &      86.4 \\
jug         & 84.6 &  98.8 &        91.8 &  99.6 &        86.6 &  82.7 &        \textbf{95.2} &  99.6 &        93.3 &      94.5 \\
bowl        & 80.4 &  78.7 &        83.6 &  80.2 &        \textbf{85.5} &  78.1 &        82.9 &  80.2 &        83.9 &      75.0 \\
\midrule
MEAN        & 88.5 &  \textbf{96.7} &        92.9 &  \textbf{96.7} &        91.5 &  92.0 &        \textbf{93.6} &  96.6 &        91.6 &      91.2 \\
\bottomrule

\end{tabular}
\captionof{table}[Results of the Standard Pipeline]{Accuracy object tracking pipeline. The upper table shows position and orientation individually. The lower table shows the area under the ADD-S curve (AUC) and $<$2cm metric.}
\label{tab:standard_results}
\end{minipage}
\vspace{-0.5cm}
\end{figure*}

\section{Experiments}
\label{sec:experiments}

\subsection{Simulated Robot Environment}
\label{ssec:simulation}

The environment consists of the Baxter robot with graspable objects on a table as shown in Figure~\ref{fig:rai_interface} right. A virtual depth camera is added on Baxter's head display to generate first-person view images (RGB-D).

The simulated environment is created using the \textit{RAI}\footnote{\url{https://github.com/MarcToussaint/rai}} interface. RAI includes a physics-simulated environment as well as a robot motion optimization solver for k-Order Motion Optimization (KOMO) problem\cite{toussaint2014komo}. The interface provides simple functionality to define motion optimization problems, by specifying the list of optimization objectives that represent cost terms or in-/equality constraints.
% The physics in the simulation are computed by PhysX and bullet.

% The resolution was chosen to 640x480 pixels, the focal length to 0.895 meter and the depth range is [0.4 meter, 7.0 meter].

\subsection{Mask R-CNN Transfer Learning on a Synthetic Dataset}

We use transfer learning to tune the Mask R-CNN to ensure the detection of custom objects. We collect images of the simulated robot environment including the following objects: a cube, a sphere, a toy, a teapot, a cup, a jug, and a bowl, as shown in Figure~\ref{fig:dataset_objects}. 

The simulated images are not like real images as they do not include noise, shadows, irregular lighting conditions, or texture. The images are augmented during training to ensure the network generalizes towards real or non-perfect images. Augmentations include flipping the image, affine transformation, light contrast, blur/sharpen, and color modifications. The dataset is also collected with arbitrary robot arm joint positions included in the image so that Mask R-CNN learns to neglect robot arms during detection. 

We obtained detection rates from the Mask R-CNN ranging from 92\% to 95\% with an error rate between 0.16\% to 0.74\% based on different training settings such as the number of epochs (10 to 80 epochs) and the dataset size (107 to 50K samples). The detection rate is defined as the number of classified objects with respect to the number of ground truth objects. The error rate is defined as the number of wrong class predictions with respect to the number of classified objects. The rest consists of undetected objects, presumably simply not detected, out of sight, or just partially visible. We do not report detailed results for each setting, as its effect turned out to be marginal when comparing the results of the whole initialization pipeline. As long as the detected label is correct, a decent mask is adequate for estimating the initial pose described in Section~\ref{ssec:pose_init}.

The model used during the evaluation of the pose initialization shown in Table~\ref{tab:standard_results} was trained with unmodified images with 40 epochs. The detection rate of the Mask R-CNN was 92.73\% with an error rate of 0.16\% out of a set of 4,798 random samples. The images of the samples were not included during the Mask R-CNN training and did not contain the arms of the Baxter robot. 

\subsection{Accuracy of the Initial Object Pose Estimation}
We evaluate the accuracy of the pose estimation using the metric proposed in \cite{xiang2017posecnn}. The average distance is computed using the closest point distance of the pairwise distances between two 3D models with ground truth transformation (translation $t$, rotation $R$) and estimated transformation (translation $\hat{t}$, rotation $\hat{R}$):
\begin{align}
    \label{eq:adds}
    \text{ADD-S} &= \frac{1}{m} \sum_{x_1 \in M} \min_{x_2 \in M} || (R x_1 + t) - (\hat{R} x_2 + \hat{t}) ||
\end{align}
in a set $\mathcal{M}$ with $m$ number of points, for both symmetric and asymmetric objects. Following prior works~\cite{xiang2017posecnn, wang2019densefusion}, we report the area under the ADD-S curve (AUC) with a threshold up to 0.1m by computing the pose accuracy while increasing the threshold. Similarly, we also measure the percentage of ADD-S below a threshold of 2cm, which is the minimum tolerance for robot grasping manipulation.

Figure~\ref{fig:acc-thres_curve} and Table~\ref{tab:standard_results} show the accuracy of different object tracking methods including the method described in Section~\ref{ssec:pose_init}. The \textit{mask\_pose}(or \textit{mask} in Table~\ref{tab:standard_results}) indicates the center position of the masked point cloud with an offset (2cm in the z-axis of the camera coordinates) added to compensate the bias in the point cloud. The \textit{mask\_pose} does not include rotation which explains the low AUC, but is a good baseline when comparing translation. The \textit{mesh\_pose} shows better precision in translation compared to the \textit{mask\_pose}, nonetheless, its main role is to provide an initial rotation estimation to initialize DBOT. 

%Note that the rotation precision of the sphere is always 97\% but should be 100\% for a perfect sphere.
%The reason is assumely the sampling of the point cloud models which are used for the accuracy
%calculation.

\subsection{Accuracy of the DBOT Tracker}
We report the result of the estimated pose once the DBOT tracker is tracking the object after receiving the initial \textit{mesh\_pose}. The object pose from the DBOT is captured after 1 second and 3 seconds after initialization while the object is kept static.
Table~\ref{tab:standard_results} shows that both trackers outperform the initial \textit{mesh\_pose}. This indicates that the tracker was able to refine the pose towards the correct object pose after receiving the estimated pose.

The particle tracker refined the pose faster and is more accurate than the Gaussian tracker. An assumption is that the Gaussian tracker is less robust to inaccurate initialization. The authors of DBOT mentioned in their paper~\cite{issac2016depth} that the particle tracker is slightly more robust, but the Gaussian tracker is more precise. 
The Gaussian tracker can tolerate distortions in the input point cloud as well as occluded settings where the particle tracker is not able to track.

We compare our objects to similar objects in the YCB dataset in terms of size and form, as shown in Table~\ref{tab:dense_res}.
A direct comparison of the results to prior work in pose detection~\cite{wang2019densefusion, xiang2017posecnn} is not completely fair, due to the different
datasets used for evaluation and that we utilized simulated images. However, it justifies the feasibility of our approach and its applicability in robot manipulation.

%\begin{center}
%\begin{tabular}{ccc}
%\textbf{YCB dataset} & {} & \textbf{own dataset}\\
%wood block & $\leftrightarrow$ & cube \\
%pitcher base & $\leftrightarrow$ & jug \\
%bowl & $\leftrightarrow$ & bowl \\
%mug & $\leftrightarrow$ & cup
%\end{tabular}
%\end{center}

\begin{table}[h!]
\vspace{-0.2cm}
\centering
\setlength\tabcolsep{2pt} % default value: 6pt
\scriptsize
\begin{tabular}{lrrrr|lrrrrrrrrrrrrrrrrrrrrrrr}
\toprule
{}  & \multicolumn{2}{c}{\text{PoseCNN}} & \multicolumn{2}{c}{DenseFusion} & \multicolumn{5}{c}{Own results}\\
{}  & \multicolumn{2}{c}{\cite{xiang2017posecnn}} & \multicolumn{2}{c}{\cite{wang2019densefusion}} & {own} & \multicolumn{2}{c}{mesh} & \multicolumn{2}{c}{particle 3s}\\
 {}                 &  AUC &  $<$2cm &  AUC &  $<$2cm & class &  AUC &  $<$2cm &  AUC &  $<$2cm \\
\midrule
pitcher 
base        & 97.8 & 100.0  &  97.1 &  100.0 &       jug    & 84.6 &  98.8       &        95.2 &  99.6             \\
bowl                & 81.0 & 54.9   & 88.2  &   98.8 &       bowl   & 80.4 &  78.7   &        82.9 &  80.2                       \\
mug                 & 95.0 & 99.8   &  97.1 &  100.0 &       cup    & 92.9 & 100.0 &        95.3 &  98.9                         \\
wood block          & 87.6 & 80.2   & 89.7  &   94.6 &       cube   & 93.8 & 100.0  &        97.0 & 100.0                 \\
\midrule
MEAN                & 90.35 & 83.73   & 93.03 & 98.35   &    &   87.93 & 94.38 & 92.6 & 94.68                        \\
\bottomrule
\end{tabular}
    \caption[Results of other Approaches]{Evaluation of 6D pose (ADD-S) on YCB-Video dataset. }
    \label{tab:dense_res}
\end{table}

The mean AUC of DenseFusion for the four objects is 93.03\%. The mean AUC of the \textit{mesh\_pose}
is 87.93\% and for the particle tracker after 3s is 92.6\%.
As already mentioned, the direct comparison is not totally fair, but as an interpretation that the object tracking pipeline is robust enough to be applicable in a robot manipulation setting.

\subsection{Accuracy of the Grasp Intent Prediction}
We collected reach-and-grab motion trajectories from two users (1 male, 1 female). Each trajectory consisted of around 2$\mathtt{\sim}$5 seconds and we collected a total of 350 trajectories for training in four different environment settings. The users started with their right hand above the Leap Motion controller and reached forward to grab a target object in a specific direction (right or top) while looking at the display similar to Figures~\ref{fig:top_grab} and~\ref{fig:right_grab}. The start and termination of the trajectories were defined by a key press. 

Figure~\ref{fig:pred_res} shows the average prediction accuracy over 18 grasp episodes of one environment using trajectories excluded during training. The overall prediction accuracies are 79.4\% and 77.4\% for target object prediction and grab direction prediction. The goal object prediction accuracy reached 100\% before reaching 70\% of the episode duration, and the average accuracy for predicting the direction reached up to 89\% at termination. The low accuracy during the first 20\% of the episode resulted from the time gap between the start of recording and the start of the movement. 

The reader may note that the prediction results are not optimal and optimization of the hyperparameters can be carried out for better results. Utilization of recurrent neural networks may also help with improving the early prediction accuracy. 

\begin{figure}[t]
\centering
\includegraphics[clip,trim={0cm 0cm 0cm 1.3cm},width=0.40\textwidth]{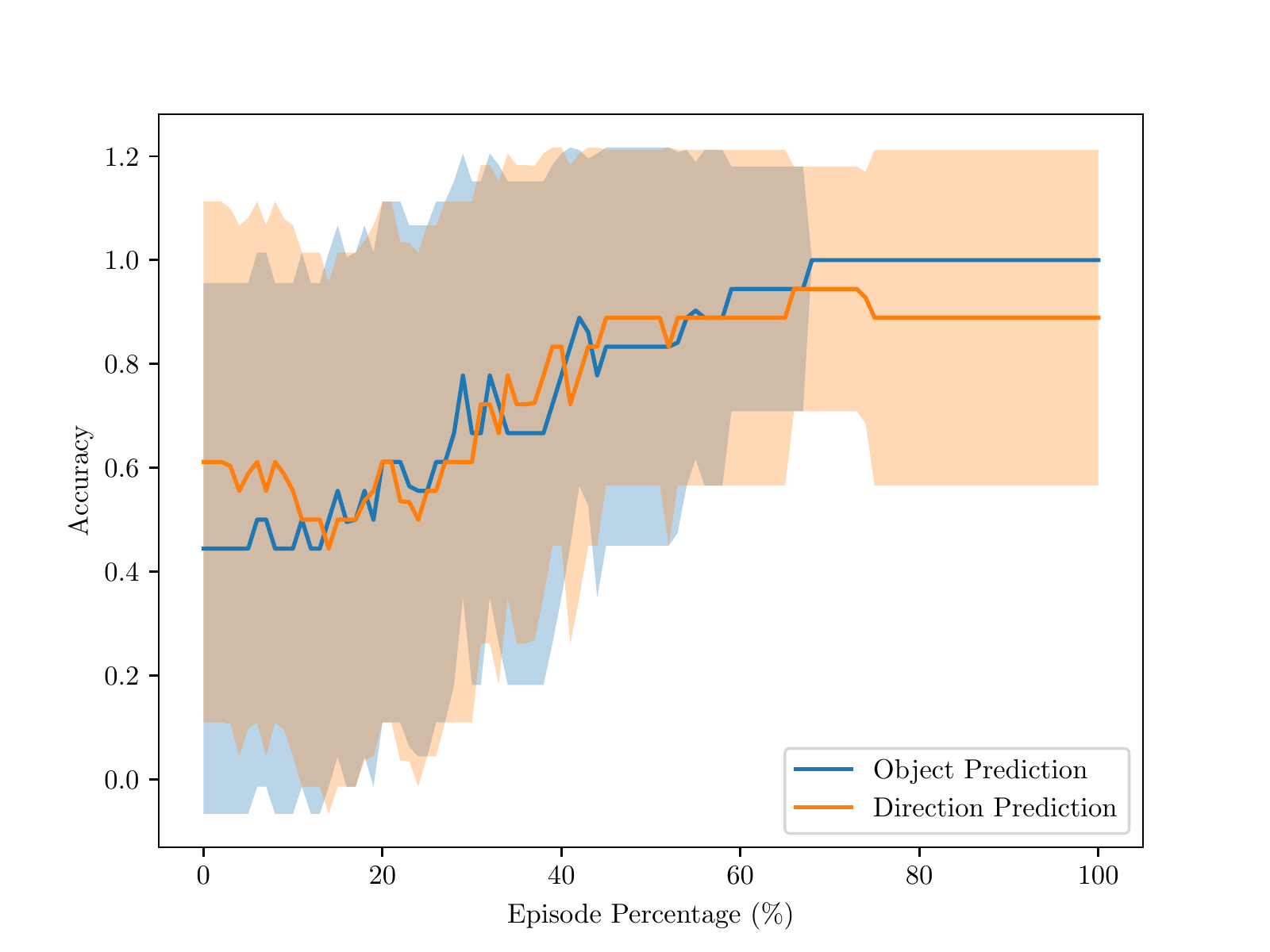}
\caption{Prediction accuracy of the grasp intention prediction averaged over the duration of the episode}
\label{fig:pred_res}
\vspace{-.3cm}
\end{figure}

\subsection{Teloperation Task Experiment Setup}

We designed a teleoperation task for manipulation to test the efficacy of the system. We hypothesize that the earlier the goal object is identified, the earlier the robot can start planning the motion, which would lead to faster task execution in a traded control setting.

% ????
%In addition, we hypothesize that we are able to identify the user's intentions despite the inconsistent Leap Motion controller with the help of an intent prediction model. 

We carried out a simulated user study by simulating different behaviors of users. We collected trajectories from three different types of virtual users 

\begin{itemize}
\item \textit{Normal} user consists of trajectories collected from human users
\item \textit{Noisy} user by injecting a Gaussian noise to the \textit{Normal} user at each time step
\item \textit{Biased} random offset over the \textit{Normal} user trajectory to simulate imperfect perception during teleoperation, e.g., recognizing the object as closer than it actually is.
\end{itemize}

The difference between a \textit{Noisy} user and a \textit{Biased} user is that the \textit{Biased} user has the same random noise over the trajectory whereas the noise in the \textit{Noisy} user changes every time step.

The task is to perform a sequence of picking motions, to grab three objects from a table. The user decides which object to grab and demonstrates the picking motion.

We assume that the poses of all objects are known by fusing the
framework presented in Section~\ref{sec:tracking}, but the robot must infer in which order the objects are grabbed. A goal object is identified when the robot predicts the same target for $t$ consecutive time steps ($t$=80). The prediction of the first $k$ time steps are neglected to reduce the prediction error in the beginning of the episode ($k$=300). 

\subsection{Evaluation of Control Modes}

We denote \textit{Early} mode as the control mode in which the robot starts to plan its grasping trajectory towards the predicted object during user demonstration.
We compare this mode with \textit{Late} mode, where the robot does not start motion planning until the user finishes the demonstration.  

The system was evaluated according to the following criteria: time taken to predict goal object (time until execution), episode duration, prediction accuracy (goal prediction, direction prediction) when the robot identified the goal. Table~\ref{tab:table_teleop} shows the results averaged over 12 episodes. The time until execution is summed up over three object grasps and the episode duration indicates the total time for picking three objects including robot motion planning time. Although it shows a compromise in the prediction accuracy, early motion planning and execution based on goal prediction resulted in shorter episode duration, as hypothesized. It is shown that it was approx. 5 seconds faster than when the robot started motion planning once the user finished the trajectory. 

The \textit{Noisy} user and the \textit{Biased} error took longer before the robot confidently identified the goal. However, there was no penalty in the prediction accuracy except in the direction prediction for the \textit{Biased} user in \textit{Early} mode. The prediction model was robust enough to tolerate the noisy settings. Overall, the results show that the proposed traded control system can improve teleoperation performance while using noisy hand gestures to control the robot.

\begin{table}
\setlength\tabcolsep{2pt} % default value: 6pt
  \begin{center}
    \begin{tabular}{c|c|r|r|r|r}
      \toprule % <-- Toprule here
       & Control& \multicolumn{3}{c}{\text{User Mode}} &\\
      Criteria & Mode &\multicolumn{1}{c}{Normal} & \multicolumn{1}{c}{Noisy} & \multicolumn{1}{c}{Biased} & \multicolumn{1}{c}{MEAN} \\
      \midrule % <-- Midrule here
      Time until&\textbf{Early}       & 9.6$\pm$1.1   & 10.3 $\pm$ 0.9 & 10.1$\pm$ 1.3 & 10.\\ 
      Execution (s)& Late & 14.6$\pm$ 1.4 & 14.7 $\pm$ 1.2 & 14.1 $\pm$ 1.5 & 14.5\\
      
      \midrule % <-- Midrule here
      
      Episode &\textbf{Early}         & 33.1 $\pm$5.6 & 32.5 $\pm$ 3.7 & 33.9$\pm$5.3 &33.2\\
      Duration (s) & Late & 38.6 $\pm$5.1 & 39.2 $\pm$ 3.1 & 38.1$\pm$5.4& 38.6\\
      
      \midrule % <-- Midrule here
      
      Object&\textbf{Early} 				& 0.86$\pm$ 0.49 & 0.89 $\pm$ 0.47 & 0.89 $\pm$ 0.62& 0.88\\
      Prediction (\%)& Late  & 0.97$\pm$ 0.28 & 0.97 $\pm$ 0.28 & 0.89 $\pm$ 0.62& 0.94\\
      
      \midrule % <-- Midrule here
      
      Direction &\textbf{Early}        & 0.94$\pm$ 0.37 & 0.97 $\pm$ 0.28 & 0.89 $\pm$ 0.62 &0.93\\
      Prediction (\%)&  Late     & 1$\pm$ 0     & 0.97 $\pm$ 0.28 & 0.97 $\pm$ 0.28& 0.98\\
      \bottomrule % <-- Bottomrule here
    \end{tabular}
  \end{center}
  \vspace{-0.3cm}
   \caption{Teleoperation results for different simulated users and control modes}
   \label{tab:table_teleop}
\end{table}

%%%%%%%%%%%%%%%%%%%%%%%%%%%%%%%%%%%%%%%%%%%%%%%%%%%%%%%%%%%%%%%%%%%%%%%%%%%%%%%%

\section{Conclusions}
\label{sec:conclusions}
We presented a teleoperation system that utilizes intuitive human grabbing hand gestures 
to perform sequential manipulation tasks. 
To mitigate the issues that arise when using hand gestures, 
%the system adopts a traded control paradigm:
 the robot autonomously generates
a grasping or retrieving motion using trajectory optimization as soon as the robot identifies the user's intention.

For the object tracking pipeline, we proposed the combination of Mask R-CNN \cite{he2017mask} and the model-based object tracker DBOT \cite{issac2016depth} for automatic initialization and object localization. 
%Results showed the successful mask detection and initial pose estimation through the proposed method, and showed the feasibility of its application in robot manipulation.
In addition, we trained a prediction model to identify the user intent from grabbing hand gestures during traded control so that the robot can start planning its trajectory in advance. 
The simulated user study indicated that using intent prediction brought down the overall task execution time. 

As the majority of our work is done in a simulated environment, limitations may arise during the application of the system in a real robot setting. We will focus on the application of the system in a real robot setting for future work. 
\vspace{-.5cm}
\addtolength{\textheight}{-1cm}   % This command serves to balance the column lengths

%%%%%%%%%%%%%%%%%%%%%%%%%%%%%%%%%%%%%%%%%%%%%%%%%%%%%%%%%%%%%%%%%%%%%%%%%%%%%%%%

\section*{Acknowledgment}

This work is partially funded by the research alliance ``System Mensch''.
The authors thank the International Max Planck Research School for Intelligent Systems (IMPRS-IS) for supporting Yoojin Oh.

%%%%%%%%%%%%%%%%%%%%%%%%%%%%%%%%%%%%%%%%%%%%%%%%%%%%%%%%%%%%%%%%%%%%%%%%%%%%%%%%

\bibliographystyle{IEEEtran}
\bibliography{IEEEabrv,bibliography,teleoperation,tim}{}

\begin{thebibliography}{10}
\providecommand{\url}[1]{#1}
\csname url@rmstyle\endcsname
\providecommand{\newblock}{\relax}
\providecommand{\bibinfo}[2]{#2}
\providecommand\BIBentrySTDinterwordspacing{\spaceskip=0pt\relax}
\providecommand\BIBentryALTinterwordstretchfactor{4}
\providecommand\BIBentryALTinterwordspacing{\spaceskip=\fontdimen2\font plus
\BIBentryALTinterwordstretchfactor\fontdimen3\font minus
  \fontdimen4\font\relax}
\providecommand\BIBforeignlanguage[2]{{%
\expandafter\ifx\csname l@#1\endcsname\relax
\typeout{** WARNING: IEEEtran.bst: No hyphenation pattern has been}%
\typeout{** loaded for the language `#1'. Using the pattern for}%
\typeout{** the default language instead.}%
\else
\language=\csname l@#1\endcsname
\fi
#2}}

\bibitem{phillips2016autonomy}
C.~Phillips-Grafflin, \emph{et~al.}, ``From autonomy to cooperative traded
  control of humanoid manipulation tasks with unreliable communication,''
  \emph{Journal of Intelligent \& Robotic Systems}, vol.~82, no. 3-4, 2016.

\bibitem{johns2016exploring}
M.~Johns, \emph{et~al.}, ``Exploring shared control in automated driving,''
  \emph{ACM/IEEE Int. Conf. on Human-Robot Interaction (HRI)}, 2016.

\bibitem{muelling2017autonomy}
K.~Muelling, \emph{et~al.}, ``Autonomy infused teleoperation with application
  to brain computer interface controlled manipulation,'' \emph{Autonomous
  Robots}, vol.~41, no.~6, 2017.

\bibitem{goil2013using}
A.~Goil, \emph{et~al.}, ``Using machine learning to blend human and robot
  controls for assisted wheelchair navigation,'' in \emph{IEEE 13th
  International Conference on Rehabilitation Robotics (ICORR)}, 2013.

\bibitem{dragan2013policy}
A.~D. Dragan and S.~S. Srinivasa, ``A policy-blending formalism for shared
  control,'' \emph{The International Journal of Robotics Research}, vol.~32,
  no.~7, 2013.

\bibitem{anderson2014experimental}
S.~J. Anderson, \emph{et~al.}, ``Experimental performance analysis of a
  homotopy-based shared autonomy framework,'' \emph{IEEE Transactions on
  Human-Machine Systems}, vol.~44, no.~2, 2014.

\bibitem{gao2014contextual}
M.~Gao, \emph{et~al.}, ``Contextual task-aware shared autonomy for assistive
  mobile robot teleoperation,'' \emph{IEEE/RSJ Int. Conf. on Intel. Robots And
  Systems (IROS)}, 2014.

\bibitem{broad2018operation}
A.~Broad, \emph{et~al.}, ``Operation and imitation under safety-aware shared
  control,'' in \emph{Workshop on the Algorithmic Foundations of Robotics},
  2018.

\bibitem{oh2020natural}
Y.~Oh, \emph{et~al.}, ``Natural gradient shared control,'' \emph{IEEE Int.
  Symp. on Robot and Human Interactive Communication (RO-MAN)}, 2020.

\bibitem{he2017mask}
K.~He, \emph{et~al.}, ``Mask r-cnn,'' \emph{IEEE Int. Conf. on Computer Vision
  (ICCV)}, 2017.

\bibitem{kofman2005teleoperation}
J.~Kofman, \emph{et~al.}, ``Teleoperation of a robot manipulator using a
  vision-based human-robot interface,'' \emph{IEEE transactions on industrial
  electronics}, vol.~52, no.~5, 2005.

\bibitem{bohren2017preliminary}
J.~Bohren and L.~L. Whitcomb, ``A preliminary study of an
  intent-recognition-based traded control architecture for high latency
  telemanipulation,'' \emph{IEEE/RSJ Int. Conf. on Intel. Robots And Systems
  (IROS)}, 2017.

\bibitem{weichert2013analysis}
F.~Weichert, \emph{et~al.}, ``Analysis of the accuracy and robustness of the
  leap motion controller,'' \emph{Sensors}, vol.~13, no.~5, 2013.

\bibitem{guna2014analysis}
J.~Guna, \emph{et~al.}, ``An analysis of the precision and reliability of the
  leap motion sensor and its suitability for static and dynamic tracking,''
  \emph{Sensors}, vol.~14, no.~2, 2014.

\bibitem{marin2016hand}
G.~Marin, \emph{et~al.}, ``Hand gesture recognition with jointly calibrated
  leap motion and depth sensor,'' \emph{Multimedia Tools and Applications},
  vol.~75, no.~22, 2016.

\bibitem{zeng2018hand}
W.~Zeng, \emph{et~al.}, ``Hand gesture recognition using leap motion via
  deterministic learning,'' \emph{Multimedia Tools and Applications}, vol.~77,
  no.~21, 2018.

\bibitem{qi2021multi}
W.~Qi, \emph{et~al.}, ``Multi-sensor guided hand gestures recognition for
  teleoperated robot using recurrent neural network,'' \emph{IEEE Robotics and
  Automation Letters}, 2021.

\bibitem{wuthrich2013probabilistic}
M.~W{\"u}thrich, \emph{et~al.}, ``Probabilistic object tracking using a range
  camera,'' \emph{IEEE/RSJ Int. Conf. on Intel. Robots And Systems (IROS)},
  2013.

\bibitem{issac2016depth}
J.~Issac, \emph{et~al.}, ``Depth-based object tracking using a robust gaussian
  filter,'' \emph{IEEE Int. Conf. Robotics And Automation (ICRA)}, 2016.

\bibitem{xiang2017posecnn}
Y.~Xiang, \emph{et~al.}, ``Posecnn: A convolutional neural network for 6d
  object pose estimation in cluttered scenes,'' \emph{arXiv preprint
  arXiv:1711.00199}, 2017.

\bibitem{wang2019densefusion}
C.~Wang, \emph{et~al.}, ``Densefusion: 6d object pose estimation by iterative
  dense fusion,'' \emph{IEEE Conf. on Computer Vision and Pattern Recognition
  (CVPR)}, 2019.

\bibitem{probreg}
\BIBentryALTinterwordspacing
{Kenta-Tanaka et al.}, ``probreg.'' [Online]. Available:
  \url{https://probreg.readthedocs.io/en/latest/}
\BIBentrySTDinterwordspacing

\bibitem{toussaint2014komo}
M.~Toussaint, ``Komo: Newton methods for k-order markov constrained motion
  problems. e-print,'' \emph{arXiv preprint arXiv:1407.0414}, 2014.

\end{thebibliography}

\end{document}